\documentclass[letterpaper, 10 pt, conference]{ieeeconf}

\IEEEoverridecommandlockouts                              
\usepackage[pdftex]{graphicx}
\usepackage{url}
\usepackage{hyperref}

\usepackage{amsmath}
\usepackage{amsfonts}
\usepackage{algorithm}
\usepackage{algorithmic}
\usepackage{bm}
\usepackage{listings}
\usepackage{color}

\lstdefinestyle{URDF}{
    language=XML,
    basicstyle=\ttfamily\scriptsize,
    showstringspaces=false, 
    keywordstyle=\color{blue},
    commentstyle=\color{green},
    stringstyle=\color{red},
    morekeywords={robot, link, joint, origin, axis, parent, child, geometry, box, size, material, color, visual, loop, predecessor, successor, coupling, ratio} 
}

\newcommand{\cg}{\mathcal{G}_C}
\newcommand{\idg}{\vec{\mathcal{G}}_D}
\newcommand{\lacg}{\mathcal{G}_A}
\newcommand{\edge}{edge}

\newcommand{\newrdf}{URDF+}
\newcommand{\el}[1]{\verb~\small{\detokenize{<#1>}}~}

\newcommand{\parent}{\lambda}
\newcommand{\motionSubspace}{\mathbf{S}}
\newcommand{\forceSubspace}{\bm{\Psi}}

\newcommand{\q}{\mathbf{q}}
\newcommand{\qd}{\dot{\mathbf{q}}}
\newcommand{\qdd}{\ddot{\mathbf{q}}}
\newcommand{\y}{\mathbf{y}}
\newcommand{\yd}{\dot{\mathbf{y}}}
\newcommand{\ydd}{\ddot{\mathbf{y}}}
\newcommand{\K}{\mathbf{K}}
\newcommand{\kk}{\mathbf{k}}

\newcommand{\g}{\mathbf{g}}

\title{\LARGE \bf \newrdf: An Enhanced URDF for Robots with Kinematic Loops}

\author{Matthew Chignoli\footnotemark$^{1}$, Jean-Jacques Slotine\footnotemark$^{1}$, Patrick M. Wensing\footnotemark$^{2}$, and Sangbae Kim\footnotemark$^{1}$
\thanks{$^{1}$Department of Mechanical Engineering, Massachusetts Institute of Technology, Cambridge, MA 02139, USA: {\tt\small chignoli@mit.edu}}
\thanks{$^{2}$Department of Aerospace and Mechanical Engineering, University of Notre Dame, Notre Dame, IN 46556, USA}}%

\begin{document}

\maketitle
\thispagestyle{empty}
\pagestyle{empty}

\begin{abstract}
Designs incorporating kinematic loops are becoming increasingly prevalent in the robotics community.
Despite the existence of dynamics algorithms to deal with the effects of such loops, many modern simulators rely on dynamics libraries that require robots to be represented as kinematic trees. 
This requirement is reflected in the de facto standard format for describing robots, the Universal Robot Description Format (URDF), which does not support kinematic loops resulting in closed chains.
This paper introduces an enhanced URDF, termed \newrdf, which addresses this key shortcoming of URDF while retaining the intuitive design philosophy and low barrier to entry that the robotics community values.
The \newrdf~keeps the elements used by URDF to describe open chains and incorporates new elements to encode loop joints.
We also offer an accompanying parser that processes the system models coming from \newrdf~so that they can be used with recursive rigid-body dynamics algorithms for closed-chain systems that group bodies into local, decoupled loops.
This parsing process is fully automated, ensuring optimal grouping of constrained bodies without requiring manual specification from the user.
We aim to advance the robotics community towards this elegant solution by developing efficient and easy-to-use software tools.
\end{abstract}


\section{Introduction}

The recursive dynamics algorithms upon which modern rigid-body dynamics (RBD) libraries are built~\cite{orin1979kinematic,featherstone1983calculation} were initially developed only for open-chain systems.
To date, these libraries~\cite{Felis2016,featherstone2014rigid,drake,pinocchioweb,todorov2012mujoco} have not implemented techniques for dealing with kinematic loops that are as simple or efficient as the original recursive algorithms.
Instead, they resort to either (i) approximating their dynamic effects or (ii) using non-recursive algorithms that scale poorly with the robot's dimension.
This lack of attention given to kinematic loops likely contributed to the decision made by the original developers of the Universal Robotic Description Format (URDF)~\cite{quigley2015programming} not to support the modeling of robots with kinematic loops.
Despite lacking such support, the URDF has become the de-facto standard format for describing robot models~\cite{tola2024understanding}.

\begin{figure}
    \centering
    \includegraphics[width=\columnwidth]{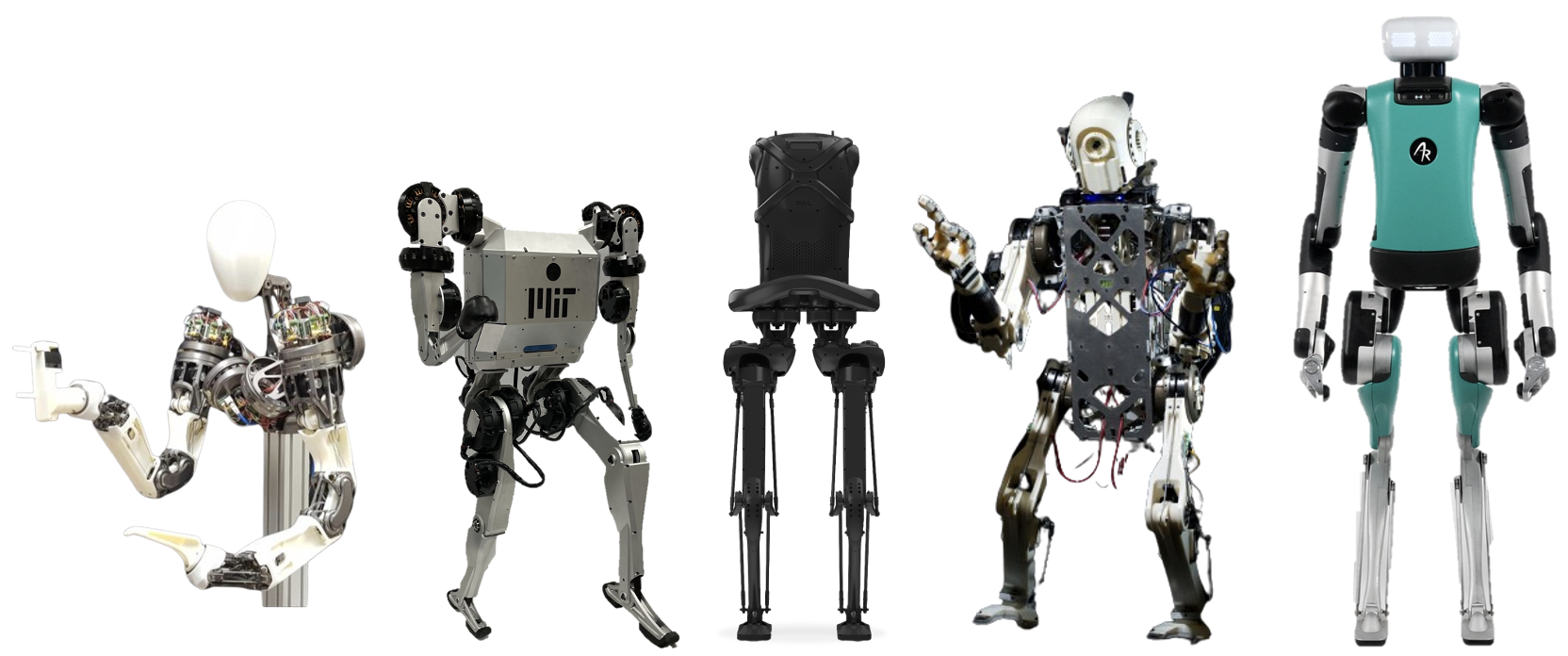}
    \vspace{-10px}
    \caption{Robots employing kinematic loops to achieve proximal actuation. Left to right: LIMS2-AMBIDEX~\cite{song2018development}, MIT Humanoid~\cite{chignoli2021humanoid}, Kangaroo~\cite{roig2022hardware}, Hermes~\cite{wang2015hermes}, Digit~\cite{digit}.}
    \label{fig:robots}
\end{figure}

With designs involving kinematic loops becoming increasingly popular (Fig.~\ref{fig:robots}) as a means to achieve proximal actuation~\cite{sim2022tello}, this shortcoming is no longer acceptable.
Designs such as parallel belt transmissions~\cite{chignoli2021humanoid}, differential drives~\cite{wang2015hermes,sim2022tello,liu2022design}, and four-bar mechanisms~\cite{apgar2018fast,liu2022design,roig2022hardware} enable high-speed limb motion while focusing the inertia of the actuators in the robot's base structure.

While the original recursive dynamics algorithms were developed for open chains, they can be adapted to systems with kinematic loops.
This was first recognized by Jain, who approached the problem through Spatial Kernel Operators (SKO)~\cite{jain1991unified} and demonstrated that systems with kinematic loops can be represented with SKO models~\cite{jain2009recursive}.
Thus, they are compatible with the original recursive algorithms~\cite{jain1990recursive}.
In our recent work~\cite{chignoli2023recursive}, we provided a self-contained derivation of these algorithms from Featherstone's perspective of propagation methods~\cite{featherstone2014rigid}, which is the prevailing perspective among modern RBD libraries~\cite{Felis2016,drake,pinocchioweb,todorov2012mujoco}.

These ``constraint-embedding" algorithms for closed chains have yet to achieve widespread proliferation.
One possible reason may be a lack of efficient, easy-to-use software tools employing these techniques.
The goal of this paper, along with our related RBD library~\cite{grbda}, is to push the robotics community toward embracing this elegant solution for dealing with a critical problem facing the field.

However, we want to emphasize that our push does not involve a paradigm shift away from the URDF.
Many commendable attempts at larger-scale shifts have been proposed~\cite{ivanou2021robot}.
For example, the Simulation Description Format (SDF)~\cite{sdf} offers many of the features requested by URDF users~\cite{tola2023understanding}, such as support for multiple robots, support for several types of sensors, and support for kinematic loops.
MuJoCo's MJCF format~\cite{todorov2012mujoco} uses a kinematic-tree-based design philosophy similar to URDF's.
The MJCF supports new and more detailed elements compared to URDF, such as sensors, actuators, constraints, and contacts.
The Asynchronous Multi-Body Framework (AMBF) Format~\cite{munawar2019real}, on the other hand, uses an entirely different modular design philosophy aimed at improving human readability and constraint handling.

Despite these efforts, a majority of the community believes that URDF will be more commonly used in the future~\cite{tola2023understanding}.
Therefore, we accept this sentiment and choose to augment, rather than replace, the URDF.
The main features of our augmented format, the \newrdf, are:
\begin{itemize}
    \item Simple additions to the original URDF data structures that allow for many more constraints to be modeled,
    \item A new parser that automatically produces models whose bodies are optimally grouped according to Jain's minimal aggregation criteria~\cite{jain2012multibody}.
\end{itemize}
When the kinematic loops of the system are ``local,'' i.e., involve a small number of bodies, the optimal grouping makes the parsed model well-suited for constraint-embedding algorithms~\cite{jain2009recursive,chignoli2023recursive}.
In cases where the loops are not local and non-recursive methods such as~\cite{carpentier2021proximal} are more efficient, the optimal parsing is still useful in providing the sparsity pattern of the constraint Jacobians.
The \newrdf~data structures and parser are implemented open-source~\cite{mit_urdf_headers,mit_urdf} as forks of the ROS URDF parser~\cite{urdf_headers,urdf}.

In this work, we also provide examples demonstrating how \newrdf~can model complicated closed-chain robots incompatible with the existing URDF format.
We emphasize that the \newrdf~retains the design philosophy of the URDF with which so many in the community are familiar.
\newrdf~files are fully backward-compatible with URDF.
RBD libraries and simulators can either (i) update their algorithms to use the fully recursive techniques of~\cite{jain2009recursive,kumar2022modular,chignoli2023recursive} or (ii) keep their existing algorithms and use the new description format to model and compute closed-loops using their existing non-recursive algorithms.

The rest of the paper is organized as follows.
Sec.~\ref{sec:background} provides background on robot modeling, the URDF, and recursive algorithms for closed-chain systems.
Sec.~\ref{sec:format} and~\ref{sec:parser} describe how \newrdf~modifies the data structures and parser of the existing URDF, respectively.
Examples of how the \newrdf~is employed are shown in Sec.~\ref{sec:examples}.
Finally, Sec.~\ref{sec:conclusion} concludes the work and discusses future steps toward achieving efficient and accessible dynamic simulation for robotics.

\section{Background} \label{sec:background}

\subsection{System Modeling}
Robotic systems are commonly modeled using graphs~\cite{featherstone2014rigid,jain2011graph}.
A graph $\mathcal{G}$ consists of a set of nodes $\mathcal{N}$ and undirected edges $\mathcal{E}$.
When the graph $\vec{\mathcal{G}}$ is directed (also called a digraph), its edges $\vec{\mathcal{E}}$ are directed from one node to another.
A graph describing a robotic system is called a connectivity graph~(CG) and has the following properties:
\begin{itemize}
    \item The nodes represent bodies.
    \item The \edge s represent joints.
    \item Exactly one root node represents a fixed base.
    \item The graph is undirected and connected.
\end{itemize}
A graph is called a topological tree when exactly one path exists between any two nodes in a graph. 
A spanning tree~(ST) of a CG is a subgraph containing all of the original CG nodes along with a set of \edge s in the original CG such that the subgraph is a topological tree.
These included \edge s constitute the set of tree joints $\mathcal{T}$.
The leftover \edge s constitute the set of loop joints $\mathcal{L}$.
Thus, for a CG with $N_B$ non-root nodes and $N_J$ \edge s, there will be $N_B$ tree joints and $N_L=N_J-N_B$ loop joints.
We will describe connectivity graphs by their bodies, tree joints, and loop joints, $\cg=\left(\mathcal{B}, \mathcal{T}\cup\mathcal{L}\right)$.

The properties of spanning trees are used to develop the ``regular numbering" convention for assigning identifying numbers to the nodes and \edge s~\cite{featherstone2014rigid}:
\begin{enumerate}
    \item Choose the \edge s to include in the ST.
    \item Assign the number 0 to the root node.
    \item Assign the remaining nodes from 1 to $N_B$ so each node has a higher number than its parent in the ST.
    \item Number the \edge s in the ST such that edge $i$ connects node $i$ to its parent.
    \item Number all remaining \edge s from $N_B+1$ to $N_J$ in any order.
\end{enumerate}
We will use the following index convention to distinguish between the different types of joints.
The indices $i$ and $j$ will be used to index tree joints and bodies (1 to $N_B$), $l$ will be used to index loops (1 to $N_L$), and $k$ identifies the loop joint that closes loop $l$ ($k=l+N_B)$.

The number of tree joint variables, $n$, and loop-closure constraints, $n^c$, are given by
\begin{equation}
    n = \sum_{i=1}^{N_B} n_i, \quad n^c = \sum_{k=N_B+1}^{N_J} n^c_k,
\end{equation}
where $n_i$ is the degrees of freedom permitted by the $i$th tree joint, and $n^c_k$ is the number of constraints imposed by the $k$th loop joint.

We now consider the particular case where the original CG is an ST.
Such a CG is called a ``kinematic tree" and corresponds to a robot free of kinematic loops.
As previously noted, the current URDF can only represent robots as kinematic trees.

\subsection{Joint Models}

\begin{figure}
    \centering
    \includegraphics[height=1.95in]{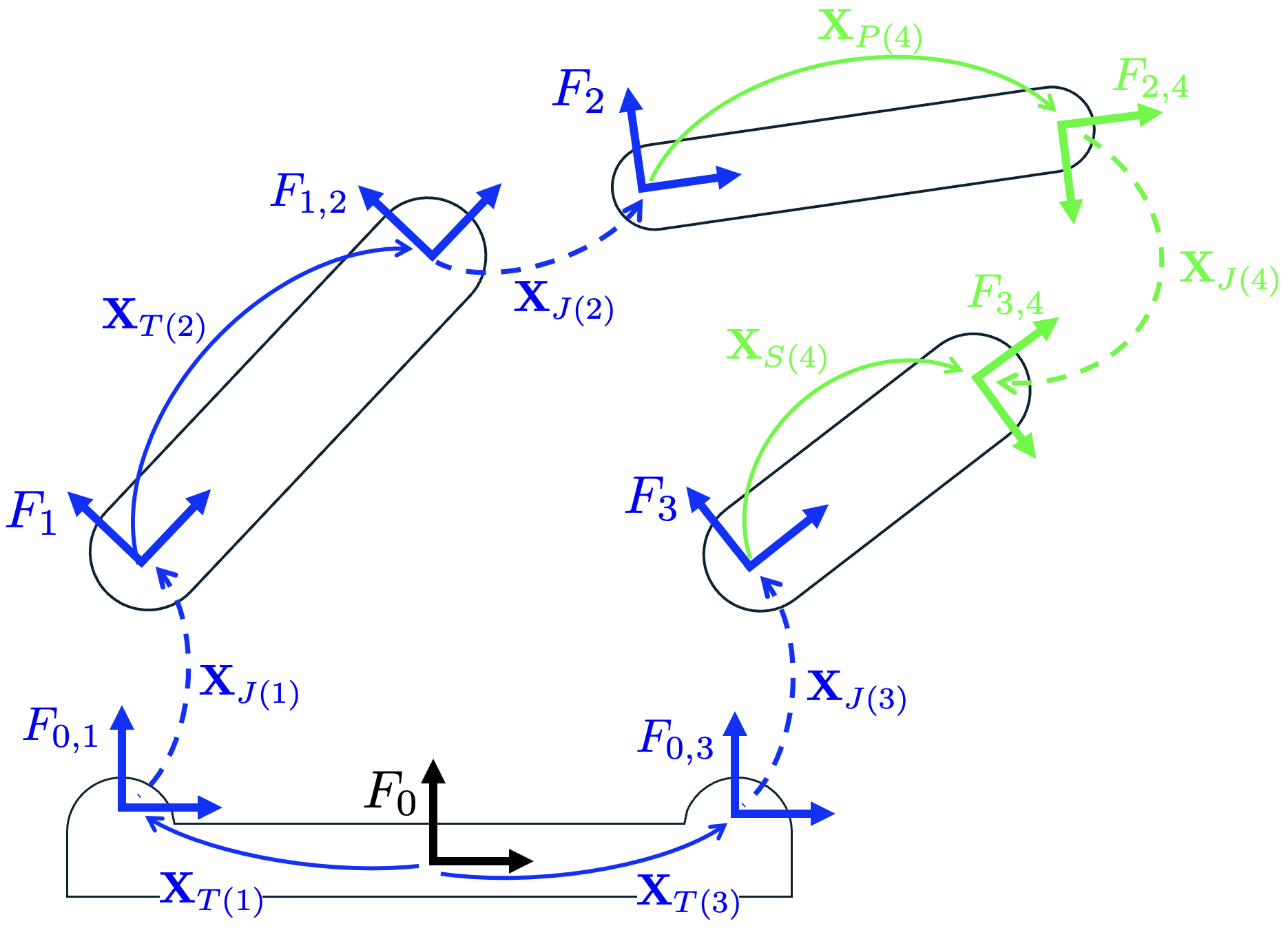}
    \caption{Exploded view of a four-bar mechanism showing its coordinate frames and the transforms between them. Tree joint quantities are shown in \textcolor{blue}{blue}, and loop joint quantities are shown in \textcolor{green}{green}.}
    \label{fig:geometry}
\end{figure}

A robot's configuration can be described by the poses of the coordinate frames attached to each of its bodies.
We use the following convention to describe a coordinate frame:~$F_{i,j}$.
The subscript $i$ denotes which body the frame is rigidly attached to.
The subscript $j$ denotes which joint the frame is associated with. 
When $i=j$, we omit $j$, leaving $F_i$.
A schematic of these coordinate frames and their relationships is shown in Fig.~\ref{fig:geometry}.

Two spatial transforms are used to transform from $F_{\parent(i)}$ to $F_i$, where $\parent(i)$ is the parent of body $i$.
The tree transform $\mathbf{X}_{T(i)}$ is a fixed transform that describes the pose of $F_{\parent(i),i}$ relative to $F_{\parent(i)}$.
This intermediate frame $F_{\parent(i),i}$ gives the location of $F_i$ when $\q_i=\mathbf{0}$.
The joint transform $\mathbf{X}_{J(i)}$ gives the pose of $F_i$ relative to $F_{\parent(i),i}$ for arbitrary $\q_i$.
The joint transform is a function of the joint position $\q_i$ and depends on the joint type.

For the $k$th loop joint, the predecessor transform $\mathbf{X}_{P(k)}$ is a fixed transform that gives the pose of $F_{p(k),k}$ relative to $F_{p(k)}$, where $p(k)$ is the predecessor body.
The successor transforms $\mathbf{X}_{S(k)}$ does the same for the successor body.
Finally, the joint transforms $\mathbf{X}_{J(k)}$ describes the transform from $F_{p(k),k}$ to $F_{s(k),k}$.

Joint models provide the information to describe the permitted relative motion between connected bodies. 
This information is captured by three quantities: $\mathbf{X}_J$, $\motionSubspace$, and $\forceSubspace$.
The motion subspace matrix $\motionSubspace_i\in\mathbb{R}^{6\times n_i}$ maps the joint velocity $\qd_i$ to the difference in 6D spatial velocity between the preceding and succeeding bodies.
Similarly, the constraint force subspace matrix $\forceSubspace_i\in\mathbb{R}^{6\times n_i^c}$ maps the constraint forces $\mathbf{f}^c_i$ to the 6D spatial force across the joint.
The joint model determines how to compute these quantities given the joint position $\q_i$.

\subsection{Loop Constraints}
The motion constraints imposed by loop joints can be expressed in either ``implicit" form
\begin{equation}
    \boldsymbol{\phi}(\q) = 0, \quad \K\qd = 0, \quad \K\qdd=\kk,
\end{equation}
or, in cases where an independent set of coordinates exists, explicit form
\begin{equation}
\q = \boldsymbol{\gamma}(\y), \quad \qd = \mathbf{G}\yd, \quad \qdd = \mathbf{G}\ydd + \g, \label{eqn:explicit_constraints}
\end{equation}
where $\q$ is the set of complete coordinates of the robot and $\y$ is the set of independent coordinates.
The constraint Jacobians and biases for the implicit and explicit constraints are given by
\[
\K = \frac{\partial \boldsymbol{\phi}(\q)}{\partial \q}, \quad \kk = -\dot{\K}\qd, \quad \mathbf{G} = \frac{\partial\boldsymbol{
\gamma}(\y)}{\partial\y}, \quad \g = \dot{\mathbf{G}}\yd.
\]

\subsection{URDF}
The main idea behind the URDF is that it encodes a kinematic tree.
The description of the entire CG is contained within the contents of the \el{robot} element.
The nodes of the graph (links of the robot) are given by the \el{link} elements.
Similarly, the \edge s of the graph (tree joints of the robot) are described by \el{joint} elements.
The contents of the \el{link} elements describe the dynamics and appearance of the respective links.
For example, the \el{inertial} child element gives the link's spatial inertia, and the \el{visual} child element provides information about its appearance (shape, size, color, etc.).
Similarly, the contents of \el{joint} elements describe kinematic constraints between links.
The \el{parent} and \el{child} elements give the links being constrained, the \el{origin} gives the tree transform $\mathbf{X}_T$, and the \verb~type~ and \el{axis} describe the joint model.
For more information on the elements and attributes comprising URDF files, see~\cite{quigley2015programming}.

\subsection{Recursive Algorithms for Closed-Chain Systems} \label{ssec:ce-algs}
The critical insight to enable recursive algorithms for closed chains, as originally demonstrated in \cite{jain2009recursive} and revisited in \cite{kumar2022modular, chignoli2023recursive}, is the transformation of non-ST connectivity graphs into ST connectivity graphs via the grouping of bodies involved in local loop constraints.
Grouping the bodies enables loop constraints to be resolved locally, i.e., only when that group of bodies is encountered during a forward or backward pass.
This local treatment avoids the need for large-scale matrix factorization.
The original presentation of the algorithms~\cite{jain2009recursive} refers to the grouping as ``constraint embedding" and the resulting groups of bodies as ``aggregate links."
In service of our propagation method-based perspective \cite{chignoli2023recursive}, we previously used the terms ``clustering" and ``clusters," respectively.
We default to the original nomenclature (constraint embedding and aggregate links) here due to less reliance on it in our subsequent development.

In modeling robots as graphs, constraint embedding involves representing multiple bodies as a single node.
Specifically, the bodies constituting an aggregate link are represented with a single node.
When all bodies are grouped in their respective aggregate nodes, the resulting connectivity graph is guaranteed to be an ST and is referred to as a loop-aggregated connectivity graph (LACG).
A LACG $\lacg$ consists of the aggregate links $\mathcal{C}$ and the collections of tree and loop joints associated with each group of bodies $\mathcal{T}_C$.
Following this process, constraints are embedded within the aggregate link and will not otherwise lead to loop constraints with other groups.
A key property of the recursive algorithms for closed chains is that their advantage over non-recursive methods diminishes as the size of the aggreate links increases~\cite{jain2012efficient,chignoli2023recursive}.
Thus, while the choice of aggreate links may be non-unique, there always exist a subset of optimal groupings.
In a thorough, graph-theoretic-based analysis of multibody system dynamics, Jain derives a criteria for minimal aggregation that, when satisfied, guarantees the model is optimally grouped~\cite{jain2012multibody}.

\section{Modifications to Description Format} \label{sec:format}

We first address the challenge of extending the description format of the URDF to accommodate kinematic loops.
In addressing this challenge, we want to be mindful of preserving the properties of the URDF that have led to its proliferation.
Specifically, the current URDF is intuitive, has a low barrier to entry, and is compatible with many software interfaces.
To preserve these properties, we ensure that \newrdf:
\begin{enumerate}
    \item Uses elements that correspond to physically meaningful properties,
    \item Requires no knowledge of constraint embedding,
    \item Minimally modifies the existing URDF,
    \item Maintains backward compatibility with the URDF.
\end{enumerate}

\newrdf~makes only three modifications to the URDF.
Two are new child elements of the \el{robot} element, and one is an optional new attribute of the \el{joint} element.
The key idea behind our modification is that we maintain all of the elements URDF uses to describe kinematic trees and instead use them to describe \textit{spanning} trees.
We then use our new elements to encode the loop joints and complete the connectivity graph.
We emphasize that the user does not have to specify the aggregate links manually. 
Aggregation takes place ``under the hood" from the user's perspective, which is important since requiring detailed knowledge of spatial kernel operators, constraint embedding, or clustering could raise the barrier to entry considerably.

\subsection{Loops}
The first element we add to the URDF to make \newrdf~is the \el{loop} element.
As noted earlier, loops refer to \edge s present in the CG but absent from the ST.
Most CGs permit multiple STs.
Thus, it is the responsibility of the user to determine which joints to declare as tree joints and which to declare as loop joints.
Some choices are more natural than others (i.e., declaring the controlled and observed joints as tree joints), but all combinations are supported by \newrdf.

The information needed to specify a \el{loop} is similar to that for a \el{joint}, although slightly more information is required.
The following template shows the full contents of the \el{loop} element:
\begin{lstlisting}[style=URDF]
<loop name="name" type="type">
    <predecessor name="name"/>
        <origin xyz="x y z" rpy="r p y"/>
    </predecessor>
    <successor name="name"/>
        <origin xyz="x y z" rpy="r p y"/>
    </successor>
    <axis xyz="x y z"/>
</loop>
\end{lstlisting}

For the $k$th loop joint, the \el{predecessor} element gives the predecessor node $p(k)$ via the \verb~name~ attribute and the predecessor transform $\mathbf{X}_{P(k)}$ via the \el{origin} child element.
The \el{successor} element provides the same information but for the successor node $s(k)$.
Like the \el{joint} element, the \verb~type~ attribute and \el{axis} child element specify the joint model.

The \el{loop} element described above contains all the information needed to formulate the implicit constraint associated with the $l$th loop.
The first step is to find the tree joints involved in the loop constraint.
This is done by finding the nearest common ancestor (NCA) between the predecessor and successor.
The ``ancestors'' of rigid body $i$ are all the rigid bodies in the ST that are on the path from the root body to body $i$.
We use $j\preceq i$ to denote that body $j$ is an ancestor of body $i$, which we emphasize is different from $j\le i$.
Thus, the NCA between predecessor and successor is given by:
\begin{equation}
    \mathrm{nca}\left((p(k),s(k)\right) = \max \left\{  i \, \vert \, i \preceq p(k), \, i \preceq s(k) \right\}.
\end{equation}

The tree joints involved in the $l$th loop are those encountered on the path from the NCA to the predecessor and those encountered from the NCA to the successor.
These sets of joints are the tree joints associated with the bodies in the path subchains $\nu_{p(k)}$ and $\nu_{s(k)}$, respectively.
These path subchains are defined by,
\begin{equation}
\begin{split}
    \nu_{p(k)} = p(k) \cup \{ i \, \vert \, \mathrm{nca}\left((p(k),s(k)\right) \prec i, \, i \preceq p(k) \}, \\
    \nu_{s(k)} = s(k) \cup \{ i \, \vert \, \mathrm{nca}\left((p(k),s(k)\right) \prec i, \, i \preceq s(k) \}.
\end{split}
\end{equation}
The standard definition of $\mathbf{K}_l$ has its $j$th block column as~\cite{featherstone2014rigid}
\begin{equation}
    \mathbf{K}_{lj} = \epsilon_{lj}\forceSubspace_k^\top\motionSubspace_j, \label{eqn:Kl}
\end{equation}
where $\epsilon_{lj}$ is $-1$ for $j\in\nu_{p(k)}$, $1$ for $j\in\nu_{s(k)}$, and $0$ for all other $j$.
For conciseness, we omit from~\eqref{eqn:Kl} the spatial transforms that ensure $\forceSubspace_k$ and $\motionSubspace_j$ are expressed in the same frame.
Recall that since the joint model is known for loop $l$, $\forceSubspace_k$ is known, and since the joint models are known for all spanning joints, all $\motionSubspace_j$ are known.

To prepare for later constraint grouping operations, we will remove all columns where $\epsilon_{lj}=0$ so that $\mathbf{K}_l$ has $n^c_k$ rows and $n_\nu$ columns instead of $n$ columns, where 
\begin{equation}
    n_\nu = \sum_{i\in\nu_p} n_i + \sum_{i\in\nu_s} n_i.
\end{equation}
The algorithmic steps for computing~\eqref{eqn:Kl} are given in Algorithm 8.4 of~\cite{featherstone2014rigid}.

\subsection{Coupling Constraints}
Joints encode kinematics constraints on the relative motion between connected bodies.
While joints capture many motion constraints encountered in robotics, they do not capture another popular type of constraint: coupling constraints.
Coupling constraints enforce relationships between joint states $\q$ rather than between motions of rigid bodies.
In other words, they \textit{couple} the constraint imposed by one joint with the constraint imposed by a different joint.
Therefore, they cannot be described by conventional joint transforms, motion subspaces, and constraint force subspaces. 
We instead make a new element, \el{coupling}, to describe such constraints. 

For now, we restrict the class of possible coupling constraints to linear relationships between the positions of tree joints.
The following template shows the full contents of the \el{coupling} element:
\begin{lstlisting}[style=URDF]
<coupling name="name">
    <predecessor name="name"/>
    <successor name="name"/>
    <ratio value="value"/>
</coupling>
\end{lstlisting}
For the coupling constraint represented as the $k$th loop joint, the \el{predecessor} and \el{successor} elements again give $p(k)$ and $s(k)$.
The coupling constraint also depends on the NCA of the predecessor and successor nodes. 
Specifically, the \el{ratio} element gives the ratio between (i) the position of the predecessor tree joint relative to the NCA and (ii) the rotation of the successor tree joint relative to the NCA\footnote{The \el{joint-mimic} element that exists for the conventional URDF is a specific case of \el{coupling} where the NCA of the predecessor and successor is also the parent of both bodies.},
\begin{equation}
    \sum_{i\in\nu_{p(k)}}\mathbf{q}_i = \eta_k \sum_{j\in\nu_{s(k)}}\mathbf{q}_j \label{eqn:coupling}
\end{equation}
Note that~\eqref{eqn:coupling} requires all of the tree joints associated with the bodies in $\nu_p$ and $\nu_s$ to have the same number of degrees of freedom and to encode the same type of motion (i.e., rotation or translation).

Unlike \el{loop} constraints, which tend to be most naturally represented as implicit constraints, coupling constraints are most naturally represented as explicit constraints, where the explicit constraint Jacobian $\mathbf{G}_k$ is a function of only $\eta_k$.
A geared transmission is the most common example of a coupling constraint, although many variations exist in robotics.

\subsection{Independent Coordinates}
If the user wishes to represent any of the loop constraints in explicit form, they must also specify a set of independent generalized velocity coordinates $\yd$.
We remark that the recursive algorithms for closed-chain systems only require the loop constraint Jacobians and biases, $\mathbf{G}_k$ and $\g_k$, to be expressed explicitly.
They do not rely on explicit constraint definitions of the form $\q=\boldsymbol{\gamma}(\y)$.
Therefore, even in cases where loop constraints are formulated as implicit (e.g.,~\eqref{eqn:Kl}), equivalent explicit constraint Jacobians and biases can be systematically derived~\cite{kumar2022modular}.

The choice of independent coordinates is non-unique, so the final modification we make to the URDF is an optional attribute allowing the user to specify which tree joints should be included in the independent coordinates and which should not.
We name this attribute \verb~independent~ and show a template for its usage here:
\begin{lstlisting}[style=URDF]
<joint name="name" type="type" independent="bool">
    <origin xyz="x y z" rpy="r p y"/>
    <axis xyz="x y z"/>
    <parent name="name"/>
    <child name="name"/>
</joint>
\end{lstlisting}
Since the attribute is optional, the model will successfully parse if omitted.
However, in that case, the user will be restricted to describing its configuration using valid sets of ST coordinates.
By making the attribute optional, we satisfy our goal of not requiring the user to have knowledge of constraint embedding.

The \newrdf~method of determining independent coordinates restricts the independent coordinates to be a subset of the complete spanning coordinates ($\yd\subseteq\qd$), even though the recursive algorithms for closed-chain systems permit alternative choices.
Furthermore, we note that for a model with $n$ degrees of freedom, the number of independent coordinates is
$    n^i = n - \sum_{l=1}^{N_L} \mathrm{rank}\left(\K_l\right)$.
To deal with cases where the user specifies an incompatible number, we have ensured that the parser will detect the incompatibbility and throw an error.

\section{Parser} \label{sec:parser}

The new data structures encoded by the \newrdf~necessitate an accompanying new parser.
The parser for the original URDF processed the elements encoding nodes and \edge s and produced a kinematic tree.
However, because the \newrdf~supports closed chains, the new parser must perform additional processing to produce a loop-aggregated connectivity graph.
This parser, specifically its automation of Jain's constraint embedding strategy~\cite{jain2012multibody}, constitutes the second contribution of this work.

\subsection{Relation to Strongly Connected Components}
Jain's constraint embedding strategy~\cite{jain2012multibody} requires the user to "identify the smallest subtree that contains [the predecessor and successor bodies of a loop constraint]."
This identification is not trivial and, especially in cases of nested or overlapping loops, may require background knowledge of constraint embedding - which we aim to avoid.
Therefore, we propose a parser that automatically identifies the minimal aggregation links.
We do this by using the concept of strongly connected components in a directed graph.
First, consider the physical interpretation of the minimal aggregation criteria: an aggregate link $\mathcal{C}$ is minimally aggregated if and only if for every pair of bodies $(B_i,B_j)\in\mathcal{C}$, a valid motion of $B_i$ cannot be computed without simultaneously computing the motion of $B_j$, and vice versa.
This condition parallels the definition of a Strongly Connected Component (SCC) in graph theory~\cite{knuth1997art}: a strongly connected component of a digraph $\vec{\mathcal{G}}=(\mathcal{N},\vec{\mathcal{E}})$ is a maximal set of vertices $\mathcal{V}\subset\mathcal{N}$ such that for all $V_i, V_j\in\mathcal{V}$, $V_i$ is reachable from $V_j$ and $V_j$ is reachable from $V_i$.

We base our \newrdf~parser on this parallel.
Specifically, our parser consists of three steps, shown in Alg.~\ref{alg:parser}.
We first create the CG by reading the URDF file, which is similar to the original URDF Parser and shown in Alg.~\ref{alg:cg}.
Next, we build a directed graph where the SCCs in the digraph correspond to minimally aggregated links.
We call this digraph a ``constraint dependency digraph."
Finally, we use a standard SCC algorithm~\cite{knuth1997art} to extract the SCCs from the constraint dependency digraph (CDD) and, therefore, form the LACG.

\begin{algorithm}
\caption{\newrdf~Parser}
\begin{algorithmic}[1]
\REQUIRE \verb!robot.urdf!
\STATE $\cg = \mathrm{connectivityGraphFromUrdf}\left(\verb!robot.urdf!\right)$
\STATE $\idg = \mathrm{constraintDependencyDigraphFromCG}\left(\cg\right)$
\STATE $\lacg = \mathrm{extractStronglyConnectedComponents}\left(\idg\right)$
\RETURN $\cg,\lacg$
\end{algorithmic} \label{alg:parser}
\end{algorithm}

\begin{algorithm}
\caption{connectivityGraphFromUrdf}
\begin{algorithmic}[1]
\REQUIRE \verb!robot.urdf!
\STATE $\mathcal{B}=\{\}$, $\mathcal{T}=\{\}$, $\mathcal{L}=\{\}$
\FOR{every \el{link} in \verb~robot.urdf~}
    \STATE Create body $B$ from \el{link}
    \STATE $\mathcal{B}\leftarrow\mathcal{B}\cup B$
\ENDFOR
\FOR{every \el{joint} in \verb~robot.urdf~}
    \STATE Build tree joint $T$ from \el{joint}
    \STATE $\mathcal{T}\leftarrow \mathcal{T}\cup T$
\ENDFOR
\FOR{every \el{loop} and \el{coupling} in \newrdf}
    \STATE Build loop joint $L$ from \el{loop} or \el{coupling}
    \STATE $\mathcal{L}\leftarrow \mathcal{L}\cup L$
\ENDFOR
\RETURN $\cg\leftarrow\left(\mathcal{B},\mathcal{T}\cup\mathcal{L}\right)$
\end{algorithmic} \label{alg:cg}
\end{algorithm}

\subsection{Constraint Dependency Digraph}
\begin{figure}[h]
    \centering
    \includegraphics[width=\columnwidth]{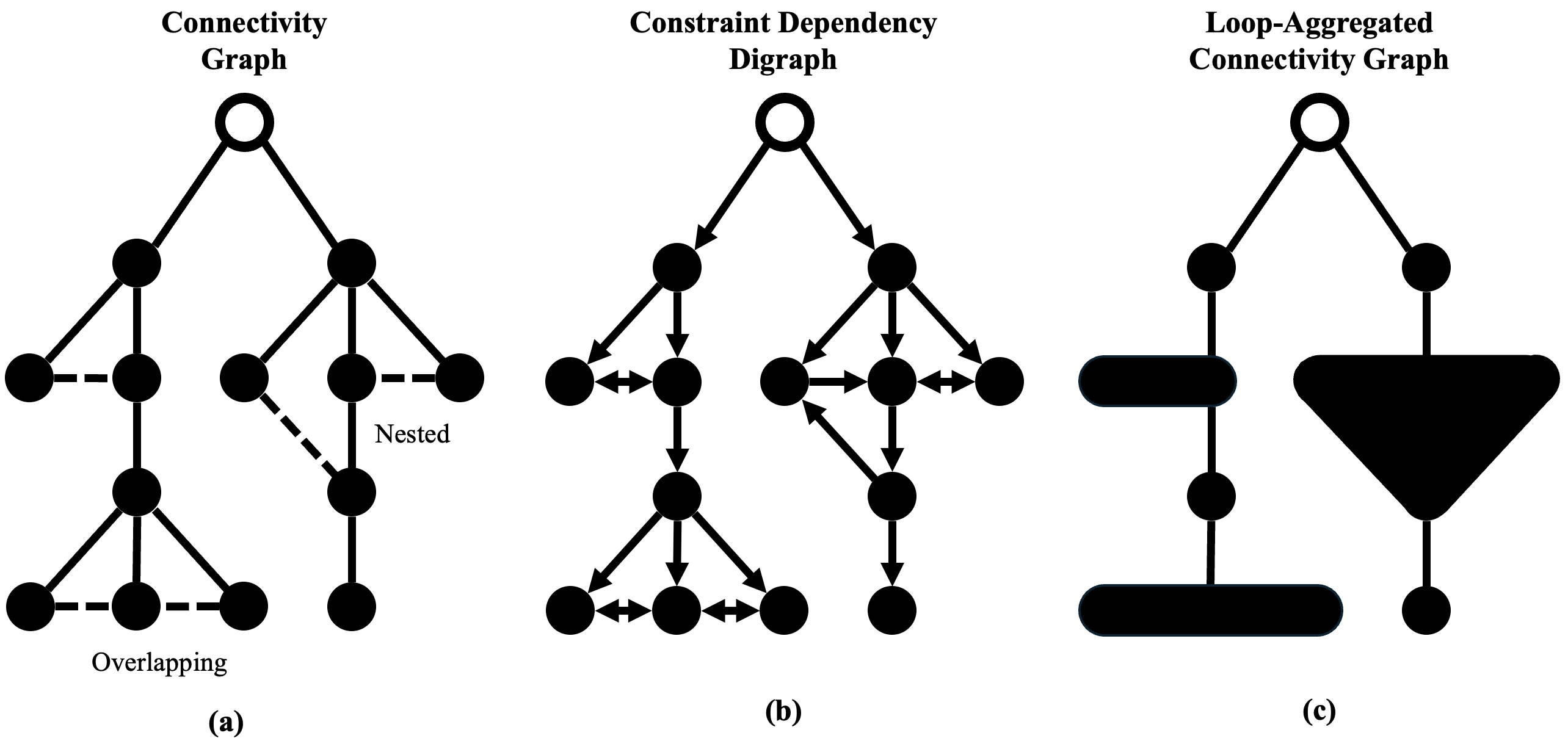}
    \caption{Illustrative connectivity graphs for a system with kinematic loops resulting in multiple aggregate links.}
    \label{fig:full_robot}
\end{figure}

Forming a constraint sub-group is not as simple as concatenating the $\nu_p$ and $\nu_s$ for a given loop joint.
For example, consider the cases of ``nested" and ``overlapping" loops, shown in Fig.~\ref{fig:full_robot}(a).
A nested loop occurs when the predecessor or successor of a loop joint $l$ is in the path subchain $\nu_{p(k')}$ or $\nu_{s(k')}$ of another loop joint $l'$.
Overlapping loops occur when a single body is the predecessor or succesor of multiple loop joints $l$ and $l'$.
In both of these cases, the motions of the bodies in $\nu_{p(k)}$ and $\nu_{s(k)}$ must be simultaneously computed with those of the bodies in $\nu_{p(k')}$ and $\nu_{s(k')}$.

Dealing with these cases, therefore, requires extra steps.
The idea behind the CDD is to use the path subchains, which are easy to find, to construct a directed graph and then allow the reachability between nodes in the digraph to determine the aggregation.
The aggregation links emerging from applying SCC decomposition to the CDD are guaranteed to satisfy the minimal aggregation criteria.
The steps for building the CDD are given by Alg.~\ref{alg:cdd} as well as depicted in Fig.~\ref{fig:full_robot}.

The constraint dependency digraph contains all the nodes in the original connectivity graph.
Every tree joint \edge~in the original spanning tree is then added to the digraph as a directed \edge~from its parent to its child. 
On the other hand, every loop joint leads to two directed \edge s in the constraint dependency digraph.
These two edges capture the dependence between the motion of the predecessor and the successor subchain and the motion of the successor and the predecessor subchain.
The first directed \edge~goes from the predecessor to the lowest numbered node in $\nu_{s}$, and the second directed \edge~goes from the successor to the lowest numbered node in $\nu_p$.
Thus, the constraint dependency digraph has $N_B$ nodes and $N_J + N_L$ directed \edge s.

\begin{algorithm}
\caption{constraintDependencyDigraphFromCG}
\begin{algorithmic}[1]
\REQUIRE $\left(\mathcal{B},\mathcal{T},\mathcal{L}\right)$
\STATE $\mathcal{N}=\mathcal{B}$, $\vec{\mathcal{E}}=\{\}$
\FOR{every tree joint $T_i\in \mathcal{T}$}
    \STATE Create directed edge $\vec{E}$ from $B_{\parent(i)}$ to $B_i$
    \STATE $\vec{\mathcal{E}}\leftarrow\vec{\mathcal{E}}\cup\vec{E}_i$
\ENDFOR
\FOR{every loop joint $L_i\in \mathcal{L}$}
    \STATE $B_{nca} = \mathrm{nca}\left((p(L_i),s(L_i)\right)$
    \STATE $\nu_p = \mathrm{pathSubchain}\left(p(L_i),B_{nca}\right)$
    \STATE $\nu_s = \mathrm{pathSubchain}\left(s(L_i),B_{nca}\right)$
    \STATE Create directed edge $\vec{E}_i$ from $p(L_i)$ to $\min \nu_s$
    \STATE Create directed edge $\vec{E}_j$ from $s(L_i)$ to $\min \nu_p$
    \STATE $\vec{\mathcal{E}}\leftarrow\vec{\mathcal{E}}\cup\{\vec{E}_i,\vec{E}_j\}$
\ENDFOR
\RETURN $\idg\leftarrow\left(\mathcal{N},\vec{\mathcal{E}}\right)$
\end{algorithmic} \label{alg:cdd}
\end{algorithm}

\begin{algorithm}
\caption{pathSubchain}
\begin{algorithmic}[1]
\REQUIRE $B_i,B_j$
\STATE $\nu = \{\}$
\WHILE{$B_i \neq B_j$}
    \STATE $\nu\leftarrow\nu\cup B_i$
    \STATE $B_i\leftarrow \parent(B_i)$
\ENDWHILE
\RETURN $\nu$
\end{algorithmic}
\end{algorithm}

\subsection{Extracting Sub-Groups}
Standard algorithms in computer science exist for automatically decomposing graphs in SCCs~\cite{knuth1997art}.
The algorithm we use involves two depth-first searches, one on the CDD and another on its reverse digraph.
The reverse digraph is formed by flipping the direction of every \edge~in the original digraph. 
Upon sorting the sub-groups, the parser stores the corresponding constraints $\K_l$ and $\mathbf{G}_l$ with the correct sub-groups, and the newly formed LACG is ready to be used in recursive dynamics algorithms.
Note that even though our parser gives an optimal sub-grouping, some of them may still be large due to the robot's morphology.
In these cases, sparsity-exploiting algorithms (e.g.,~\cite{featherstone2010exploiting,carpentier2021proximal}) may outperform recursive ones.
The \newrdf~data structures and parser are still useful because in the process of parsing the model, the original spanning tree is stored, and so too are the loop constraints $\mathbf{K}$ and $\mathbf{G}$, along with their sparsity patterns.

\section{Examples} \label{sec:examples}

We provide some illustrative examples to demonstrate how to use the new features of \newrdf~and how to apply them to model different types of closed-chain mechanisms.
Note that for the sake of space, we only include the portions of the \newrdf~that are needed to communicate how to use the new features.
Therefore, some examples might be ``underspecified" and not directly usable by an RBD library.

\subsection{LIMS2-AMBIDEX Wrist}

\begin{figure}[ht]
    \centering
    \includegraphics[width=\columnwidth]{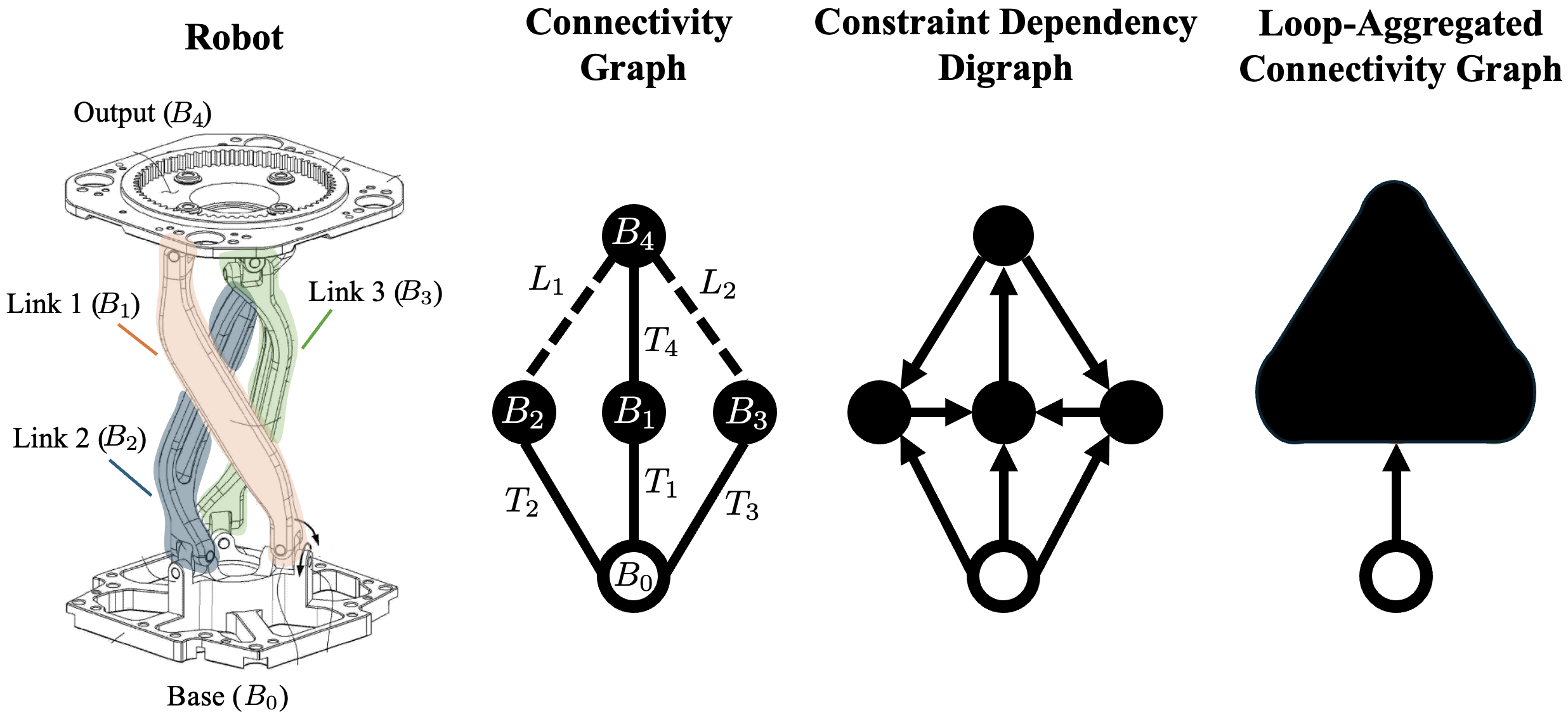}
    \caption{Schematic of the 2-DOF wrist joint for the LIMS2-AMBIDEX robot~\cite{song2018development} and its corresponding connectivity graphs.}
    \label{fig:wrist}
\end{figure}

We first provide an example using the \el{loop} element, applying it to a 2-DOF virtual rolling joint for wrist pitch/roll motion~\cite{song2018development}.
A picture of the joint is shown in Fig.~\ref{fig:wrist}.
The clever design of the mechanism allows it to emulate the pure rolling contact of two spheres while maintaining a wide range of motion free of singular poses.
The connecting rods create kinematic loops that prevent the mechanism from being accurately modeled by URDF.

The abridged \newrdf~for this mechanism is:
\begin{lstlisting}[style=URDF]
<link name="Base"/>
<link name="Link 1"/>
<link name="Link 2"/>
<link name="Link 3"/>
<link name="Output"/>
<joint type="universal" independent="true">
    <parent name="Base"/>
    <child name="Link 1"/>
</joint>
<joint type="universal" independent="false">
    <parent name="Base"/>
    <child name="Link 2"/>
</joint>
<joint type="universal" independent="false">
    <parent name="Base"/>
    <child name="Link 3"/>
</joint>
<joint type="universal" independent="false">
    <parent name="Link 1"/>
    <child name="Output"/>
</joint>
<loop type="universal">
    <predecessor name="Link 2"/>
    <successor name="Output"/>
</loop>
<loop type="universal">
    <predecessor name="Link 3"/>
    <successor name="Output"/>
</loop>
\end{lstlisting}

The connectivity graph, the constraint dependency digraph, and the loop-aggregated connectivity graph for this \newrdf~are also shown in Fig.~\ref{fig:wrist}.
Observe in the constraint dependency digraph that the connecting rods and output body are all reachable from one another.
Thus, they constitute an SCC.
Following the parsing rules from Sec.~\ref{sec:parser}, the implicit constraint Jacobian for the mechanism is
\begin{equation}
    \mathbf{K} =  \begin{bmatrix} \mathbf{K}_1 \\ \mathbf{K}_2 \end{bmatrix} = \begin{bmatrix} \forceSubspace_1^\top\motionSubspace_1 & \forceSubspace_1^\top\motionSubspace_2 & \mathbf{0} & \forceSubspace_1^\top\motionSubspace_4 \\ \forceSubspace_2^\top\motionSubspace_1  & \mathbf{0} & \forceSubspace_2^\top\motionSubspace_3 & \forceSubspace_2^\top\motionSubspace_4 \end{bmatrix}
\end{equation}
Since the total number of degrees of freedom of the tree joints is 8, and the rank of the constraint matrix is 6, the mechanism has the expected number of independent degrees of freedom: 2.

\subsection{Parallel Belt Transmission}

\begin{figure}[ht]
    \centering
    \includegraphics[width=\columnwidth]{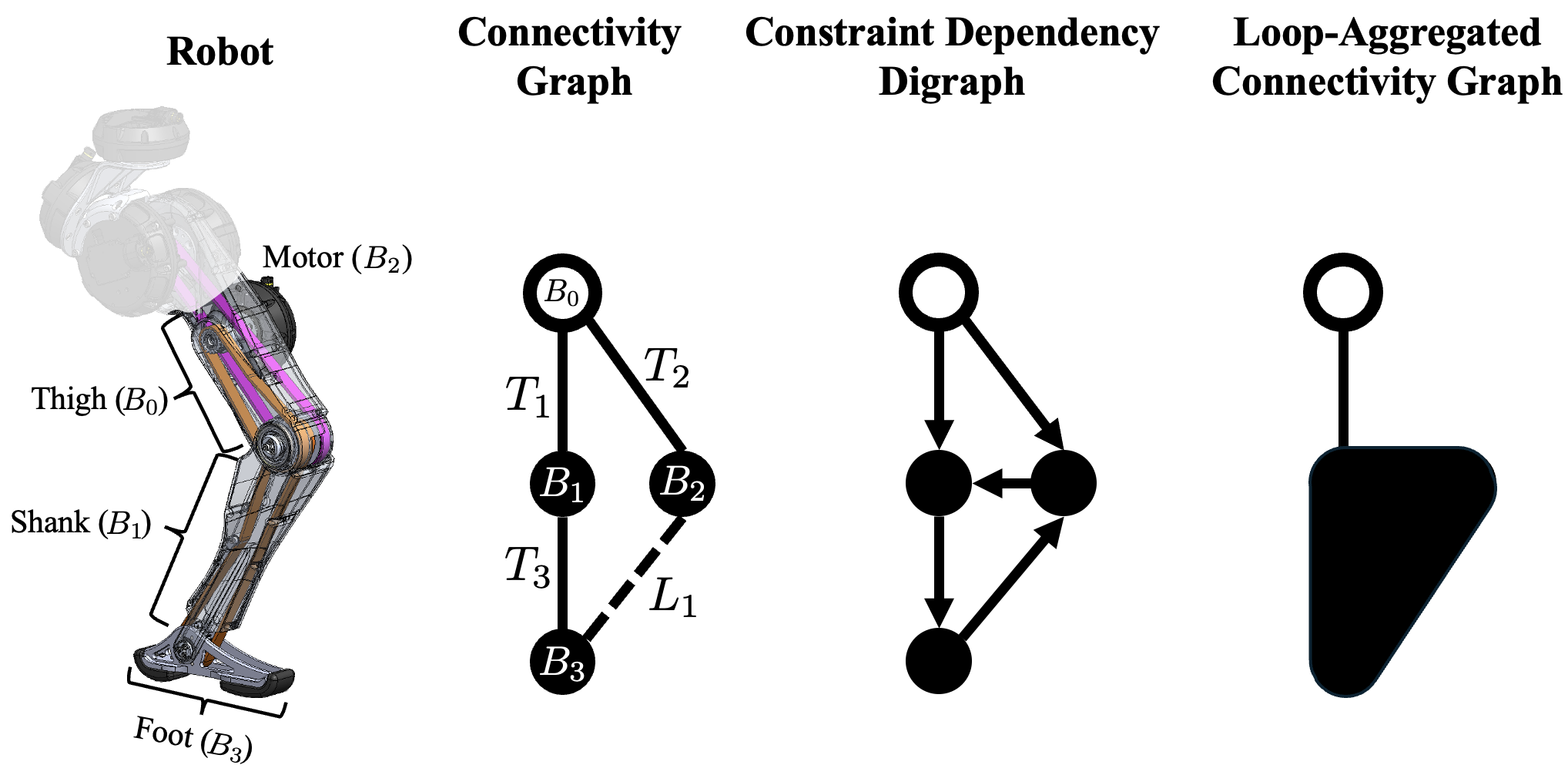}
    \caption{CAD view of the parallel belt transmission in the MIT Humanoid~\cite{chignoli2021humanoid} and its corresponding connectivity graphs.}
    \label{fig:belt}
\end{figure}

We also provide an example of a coupled constraint as part of a parallel belt transmission~\cite{chignoli2021humanoid}, shown in Fig.~\ref{fig:belt}.
The simplest example of a coupled constraint is a geared transmission.
This is easily modeled by \newrdf.
However, there are also simple changes to the conventional recursive algorithms that can approximately account for the effects of geared motors~\cite{paul1981robot}.
For a more complicated design like a parallel belt transmission where the NCA of the predecessor and successor is not also the parent of the predecessor and successor, the constraint gets more complicated, and the geared motor approximation technique fails to generalize.

The abridged \newrdf~for this mechanism is:
\begin{lstlisting}[style=URDF]
<link name="thigh"/>
<link name="shank"/>
<link name="motor"/>
<link name="foot"/>
<joint type="revolute" independent="true">
    <parent name="thigh"/>
    <child name="shank"/>
</joint>
<joint type="revolute" independent="false">
    <parent name="thigh"/>
    <child name="motor"/>
</joint>
<joint type="revolute" independent="true">
    <parent name="shank"/>
    <child name="foot"/>
</joint>
<coupling>
    <predecessor name="foot"/>
    <successor name="motor"/>
    <ratio value="eta"/>
</coupling>
\end{lstlisting}

The connectivity graph, the constraint dependency digraph, and the loop-aggregated connectivity graph for this \newrdf~are also shown in Fig.~\ref{fig:belt}.
The predecessor subchain $\nu_p$ includes the foot and shank, and the successor subchain $\nu_s$ includes only the motor.
Therefore, the explicit constraint can be expressed 
\begin{equation}
    \mathbf{q}_\mathrm{shank} + \mathbf{q}_\mathrm{foot} = \eta\mathbf{q}_\mathrm{motor}.
\end{equation}
leading to the constraint Jacobian
\begin{equation}
    \dot{\mathbf{q}} = \begin{bmatrix} \dot{\mathbf{q}}_\mathrm{shank} \\ \dot{\mathbf{q}}_\mathrm{foot} \\ \dot{\mathbf{q}}_\mathrm{motor} \end{bmatrix} = \begin{bmatrix} 1 & 0 \\ 0 & 1 \\ 1/\eta & 1/\eta \end{bmatrix} \begin{bmatrix} \dot{\mathbf{q}}_\mathrm{shank} \\ \dot{\mathbf{q}}_\mathrm{foot} \end{bmatrix} = \mathbf{G}\dot{\mathbf{y}}.
\end{equation}
\section{Conclusion} \label{sec:conclusion}

The introduction of \newrdf~represents an advancement in the ability of modern RBD libraries to seamlessly support robots with kinematic loop designs.
By enhancing the conventional URDF to include such loops while preserving its intuitive design and usability, \newrdf~addresses a critical gap in the existing standards. 
Our modifications maintain the familiar elements of URDF for describing kinematic trees and add new elements for loop joints, ensuring compatibility with recursive algorithms for closed-chain systems.
The development of an automated parser to handle the new elements and generate loop-aggregated connectivity graphs underscores the user-centric approach of \newrdf. 
This automation eliminates the need for manual specification of aggregate links, simplifying the modeling process for users and ensuring optimal performance.
Through illustrative examples and the creation of supporting tools, we demonstrate the practical benefits and feasibility of \newrdf. 
Our goal is to encourage the robotics community to adopt this enhanced format, which will facilitate the design and simulation of more complex robotic systems.
Future work will focus on adding features such as more complicated coupling constraints (e.g. differential joints), specifying arbitrary independent coordinates, and parsing models in a manner that detects when which constraints should be handled with constraint-embedding versus non-recursive alternatives such as~\cite{carpentier2021proximal} or~\cite{sathya2024constrained}.


\bibliographystyle{ieeetr}
\bibliography{references}

\end{document}